\documentclass[11pt]{article}
\UseRawInputEncoding
\usepackage[utf8]{inputenc}
\usepackage[margin=1in]{geometry}
\usepackage{graphicx}
\usepackage{amsmath,amssymb}
\usepackage{cite}
\usepackage{url}
\usepackage{algorithmic}
\usepackage{textcomp}
\usepackage{makecell}
\usepackage{adjustbox}
\usepackage{xcolor}
\usepackage{float}

\title{XAG-Net: A Cross-Slice Attention and Skip Gating Network for 2.5D Femur MRI Segmentation}

\author{
Byunghyun Ko$^{1}$, Anning Tian$^{1}$, Jeongkyu Lee$^{1}$ \\
\small $^{1}$Khoury College of Computer Sciences, Northeastern University, USA \\
\small \texttt{\{ko.by, tian.ann, jeo.lee\}@northeastern.edu}
}

\date{}

\begin{document}
\maketitle
\begin{center}
    {\small \copyright 2025 IEEE. Personal use of this material is permitted. 
    Permission from IEEE must be obtained for all other uses, 
    including reprinting, republishing, or reuse of this material for commercial purposes.}
\end{center}

\begin{abstract}
Accurate segmentation of femur structures from Magnetic Resonance Imaging (MRI) is critical for orthopedic diagnosis and surgical planning but remains challenging due to the limitations of existing 2D and 3D deep learning-based segmentation approaches. In this study, we propose XAG-Net, a novel 2.5D U-Net-based architecture that incorporates pixel-wise cross-slice attention (CSA) and skip attention gating (AG) mechanisms to enhance inter-slice contextual modeling and intra-slice feature refinement. Unlike previous CSA-based models, XAG-Net applies pixel-wise softmax attention across adjacent slices at each spatial location for fine-grained inter-slice modeling. Extensive evaluations demonstrate that XAG-Net surpasses baseline 2D, 2.5D, and 3D U-Net models in femur segmentation accuracy while maintaining computational efficiency. Ablation studies further validate the critical role of the CSA and AG modules, establishing XAG-Net as a promising framework for efficient and accurate femur MRI segmentation.
\end{abstract}
\section*{Keywords}
Medical image segmentation, Cross-slice attention, Femur MRI segmentation, 2.5D convolutional neural networks, Anisotropic volumetric data, Attention gating, Deep learning

\section{Introduction}
Precise segmentation of the femur structure from Magnetic Resonance Imaging (MRI) scans is crucial for orthopedic diagnosis and surgical planning\cite{Xie2025}. However, manual segmentation of MRI scans necessitates skilled professionals, incurs substantial labor costs, and poses risks of inaccuracies~\cite{FOURNEL2021102213}. Deep learning-based segmentation methods like U-Net often face considerable challenges as well\cite{FOURNEL2021102213, ronneberger2015unet}. Typically, 2D segmentation approaches, i.e., 2D U-Net\cite{ronneberger2015unet}, process images slice-by-slice, which offers great computational efficiency. However, 2D approaches often fail to capture contextual information along the inter-slice axis~\cite{ZHANG2022102088}. On the other hand, 3D segmentation tasks, i.e., 3D U-Net\cite{Cicek2016_3DUNet}, process images directly with volumetric data using 3D convolutions, which allows more comprehensive spatial feature extraction, but often requires higher computational resources and memory~\cite{ZHANG2022102088}.

2.5D segmentation approaches focus on segmenting objects by leveraging partial volumetric context provided by a 2.5D input\cite{ZHANG2022102088}. Unlike 2D images or 3D volumes, 2.5D images are created by stacking a series of 2D images together \cite{ZHANG2022102088}. This input shape of 2.5D approaches allows them to utilize 2D convolutions in their architecture while preserving partial 3D context. 2.5D segmentation methods are often limited by insufficient representation of inter-slice continuity and incomplete utilization of spatial context~\cite{Nikzad2024CSANet}. Many also either neglect pixel-level normalization of attention across slices or impose computationally heavy architectures, which leaves room for innovation in efficiently modeling fine-grained inter-slice relationships~\cite{Nikzad2024CSANet}. 

In this study, we propose a novel 2.5D U-Net architecture called XAG-Net and a cross-slice attention (CSA) module. Our unique CSA module employs a per-pixel softmax attention operation across three slices, which are applied to input feature maps and complemented by residual connections. The CSA module and Attention Gating (AG) blocks within XAG-Net together provide critical inter- and intra-slice contextual modeling for femur segmentation. On a femur MRI dataset, XAG-Net demonstrates state-of-the-art performance, surpassing baseline models in accuracy - with up to 12.3 percentage point gain in dice score - and boundary precision while remaining computationally efficient. An extensive combinatorial ablation study validates the complementary roles of XAG-Net’s architectural components.

\section{Related Work}
Recently, deep learning techniques have become a popular way to gain accurate segmentation results for femur segmentation. Femur segmentation offers a unique challenge due to rigid cortical boundaries with minor anatomical variations. A 3D convolutional neural network (CNN) approach has achieved a Dice similarity coefficient (DSC) score of 0.94 \cite{Deniz_2018} while a 2D U-Net approach has achieved a DSC score of 0.932 \cite{liu2025performanceanalysisdeeplearning}. However, although current approaches can reach high overlap scores, none have proposed solutions to solve the issues arising from inter‑slice discontinuities and small cortical structures that are harder to segment. 

Broadly, CSA is a mechanism that utilizes a stack of a fixed number of neighboring slices as input. Although the exact CSA mechanism varies per model, it often comprises a 1 $\times$ 1 convolution or a linear projection that produces an attention score for each slice. Many 2.5D CSA-based mechanisms like CSA-Net\cite{Nikzad2024CSANet} and CSAM\cite{Yu2024CSAM} share similar goals of allowing models to preserve some volumetric context while avoiding the high computational costs of 3D convolutions. While 3D approaches allow the model to leverage full volumetric context, it is computationally expensive and may obscure local features essential for accurate femur segmentation. Therefore, a 2.5D input structure combined with CSA modules is a promising direction for the architecture design for femur segmentation.

In contrast, AG blocks often operate within a single slice to provide intra-slice context by suppressing irrelevant background noises\cite{oktay2018attentionunetlearninglook}. These blocks are learnable gates that utilize the encoder's features and decoder's gating signal to tune each skip connection's feature map. By doing this, AG blocks suppress background clutter and highlight anatomies most relevant for segmentation \cite{oktay2018attentionunetlearninglook}. Thus, utilizing CSA for inter-slice context and AG for intra-slice refinement may offer a viable strategy for capturing both local and contextual features necessary for femur segmentation. These existing gaps and advances collectively motivate the dual CSA + AG design of XAG‑Net's architecture. 
\begin{figure*}[t]
    \centering
    \includegraphics[width=1\textwidth]{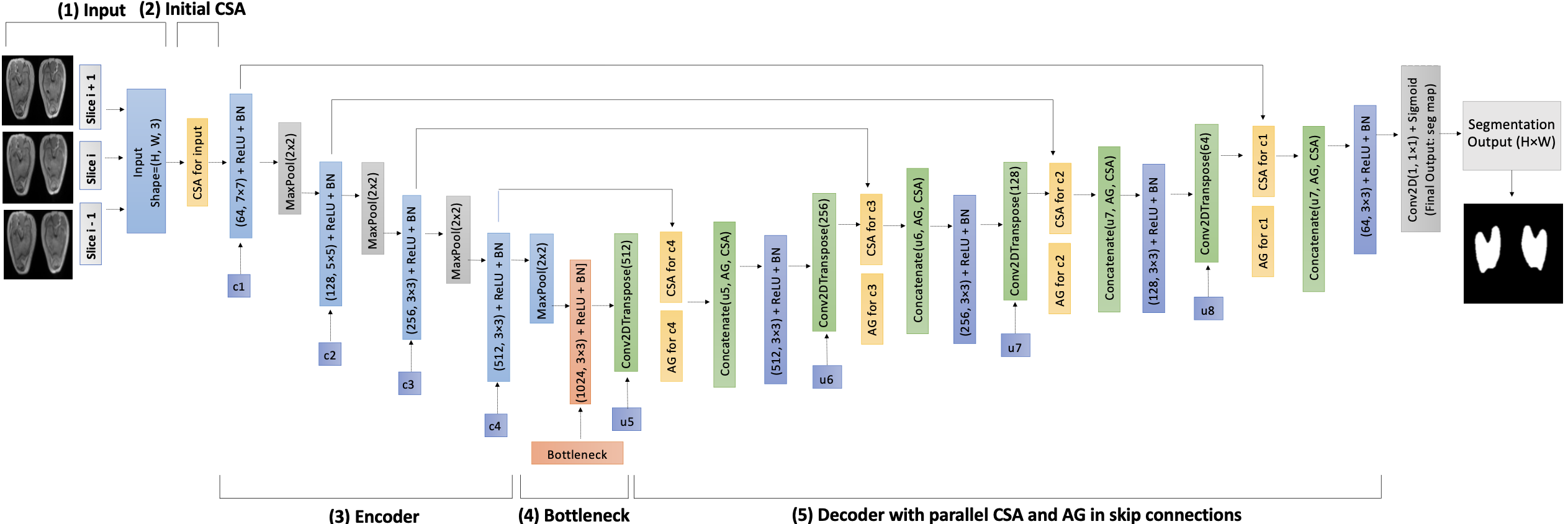}
    \caption{An architectural overview of the proposed XAG-Net}
    \label{fig:xag_net_architecture}
\end{figure*}\
\begin{figure*}[t]
    \centering
    \includegraphics[width=0.9\textwidth]{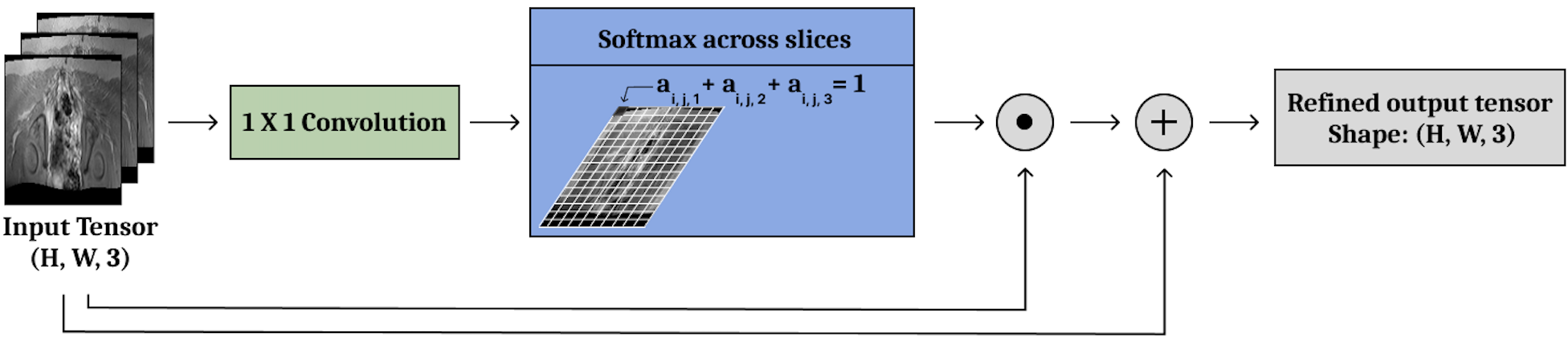}
    \caption{An overview of XAG-Net’s CSA module: The softmax operation ensures that the attention weights assigned to each pixel across the three slices sum to 1. The $\odot$ symbol represents element-wise multiplication and the $+$ symbol represents a residual connection.}
    \label{fig:csa_module}
\end{figure*}\
\section{Methods}
XAG-Net is a novel 2.5D U-Net architecture developed to
leverage inter-slice and intra-slice contextual information when
segmenting femur bone structures from MRI data. The model takes three sequential axial slices from the
MRI data and stacks them as a single input. As shown in Fig.~\ref{fig:xag_net_architecture}, the 2.5D input shape allows the model to execute efficient 2D convolutions on features that contain partial volumetric context \cite{2.5D_UNet}. XAG-Net also incorporates cross-slice attention (CSA) modules and attention gating (AG) blocks throughout its architecture, allowing the model to leverage both inter-slice contextual relationships and intra-slice feature refinement.

\subsection{Cross-Slice Attention (CSA) Module}

A novel CSA module is implemented in both the initial input stage and the skip connections of the proposed XAG-Net. Various studies have explored different approaches to incorporating cross-slice attention for MRI segmentation tasks such as CSA-Net\cite{Nikzad2024CSANet}, CSAM\cite{Yu2024CSAM}, and RSANet\cite{Zhang_2019}. These models apply variations of cross-slice attention, but none uses a pixel-wise softmax normalization across adjacent slices at each spatial location in a way that XAG-Net's CSA module does. A unique characteristic of this CSA module is that it explicitly normalizes slice contributions against each other. Fig.~\ref{fig:csa_module} provides an overview of the CSA module. 

Given an input tensor $X \in \mathbb{R}^{H \times W \times C}$, where $H$ and $W$ each represent the height and width of the spatial dimension and $C$ is the number of slices, a $1\times1$ convolution is applied to predict attention maps across the slices at each spatial location. These attention maps are then normalized via a softmax operation along the slice (channel) dimension for each pixel location as shown in equation (\ref{eq:csamodule}):
\begin{equation}
   \textit{CSA}(X) = X + \left( X \odot \textit{Softmax}\left( W(X) \right) \right)
    \label{eq:csamodule}
\end{equation}, where $W(\cdot)$ and $\odot$ represent the convolutional layer and element-wise multiplication, respectively. The residual connection provides stable learning for the model by preserving the original features. This CSA module is implemented at both the input stage and within each skip connection of XAG-Net.
\subsection{Attention Gating (AG) Block}
AG blocks, first introduced in \cite{oktay2018attentionunetlearninglook}, are implemented in the skip connections of the proposed XAG-Net. Each AG block takes as input an encoder feature map $x$ and a gating signal 
$g$, which is the corresponding decoder feature map from a deeper layer. The gating signal provides contextual guidance that helps the network focus on salient anatomical structures during feature fusion \cite{oktay2018attentionunetlearninglook}.

Given an encoder feature map $x$ and a gating signal $g$ from the decoder, the attention coefficient $\alpha$ is computed by equation (\ref{eq:agcoefficient}):
\begin{equation}
    \alpha = \sigma\left( W_{\theta} x + W_{\phi} g + b \right)
    \label{eq:agcoefficient}
\end{equation}, where $W_{\theta}$ and $W_{\phi}$ are learnable $1\times1$ convolutions, $b$ is a bias term, and $\sigma(\cdot)$ represents the sigmoid activation function. AG is computed with equation (\ref{eq:ag}):
\begin{equation}
    \textit{AG}(x) = \alpha \odot x
    \label{eq:ag}
\end{equation}, where $\odot$ represents element-wise multiplication. By dynamically adjusting skip connections based on decoder context, AG allows the model to extract highly-detailed anatomical structures \cite{oktay2018attentionunetlearninglook}.

\subsection{Architecture Design}
The architectural design of XAG-Net is an extension of a 2.5D U-Net architecture with modifications made specifically in the input and skip connections as shown in Fig.~\ref{fig:xag_net_architecture}. Like other U-Net-based models, XAG-Net follows an encoder-decoder structure along with skip connections \cite{ronneberger2015unet}. The key components of XAG-Net are as follows: 

\textit{1) Input}: Three axial femur MRI slices are stacked along the channel dimension of $(256 \times 256 \times 3)$, forming a 2.5D structure. This stack of three MRI axial slices allows the model to conduct 2D convolutions on three slices simultaneously.

\textit{2) Initial CSA}: An initial CSA is applied to refine the input before it enters the encoder. This allows the feature maps of the input to be adjusted based on the inter-slice context before entering the encoder.  

\textit{3) Encoder}: We place four down-sampling stages at the encoder. Each stage has convolutional layers with ReLU activation, batch normalization, and max pooling. The number of filters doubles at each stage as the features are passed forward. 

\textit{4) Bottleneck}: This is the deepest convolutional block with 1,024 filters. The bottleneck is situated between the encoder and the decoder. 

\textit{5) Decoder}: We place four upsampling stages at the decoder. At each stage, the upsampled feature map from the previous decoder stage is concatenated with the outputs of the CSA module and the AG block. The CSA and AG modules are implemented separately, or in a parallel manner, on the encoder features. 

Given that $U$ is the upsampled feature map from the previous decoder, $S_{enc}$ is a skip connection feature map from encoder, and $g$ is a gating signal from decoder (upsampled decoder feature), the outputted feature map $F_{concat}$ at each decoder stage is defined by equation (\ref{eq:feature_map}):
\begin{equation}
    F_{concat} = \textit{Concat}(U, \textit{CSA}(S_{enc}), \textit{AG}(S_{enc}, g))
    \label{eq:feature_map}
\end{equation}, where $\textit{Concat}(\cdot)$ represents channel-wise concatenation $\textit{CSA}(\cdot)$ denotes the cross-slice attention module, and $\textit{AG}(\cdot)$ represents the attention gating mechanism. 

This outputted feature map goes through a convolutional layer and batch normalization before being passed to the next decoder stage. At the end of the final decoder stage, a $1 \times 1$ convolution is implemented to the outputted decoder feature to the output mask of $(256 \times 1 \times 1)$. A sigmoid activation is then employed as the final step. 

\subsection{Loss Function}
The loss function utilized for the training of XAG-Net is a combined loss function comprised of boundary loss and Dice loss \cite{medvissuite}. Given $|y_{true}|$ is the sum of ground truth pixels and $|y_{pred}|$ is the sum of predicted pixels, the Dice score is computed with equation (\ref{eq:dice_score}): \begin{equation}
    \textit{Dice}(y_{true}, y_{pred}) = \frac{2|y_{true} \cap y_{pred}| + \epsilon}{|y_{true}| + |y_{pred}| + \epsilon}
    \label{eq:dice_score}
\end{equation}
, where a smoothing constant of $\epsilon = 1$ is added to both the numerator and the denominator for numerical stability. 

$L_{Dice}$, which measures the overlap between predicted and ground truth masks, is defined by equation (\ref{eq:dice_loss}): \begin{equation}
    L_{Dice}(y_{true}, y_{pred}) = 1 - \text{Dice}(y_{true}, y_{pred})
    \label{eq:dice_loss}
\end{equation}
Given $\nabla$ is the Sobel gradient operator, $L_{Boundary}$, which penalizes errors along the boundaries, is defined by equation (\ref{eq:boundary_loss}): \begin{equation}
    L_{Boundary}(y_{true}, y_{pred}) = \textit{mean}\left( \left| \nabla y_{true} - \nabla y_{pred} \right| \right)
    \label{eq:boundary_loss}
\end{equation}
Finally, the combined loss is calculated with equation (\ref{eq:combined_loss}): \begin{equation}
    L_{combined} = 0.9 \times L_{Dice} + 0.1 \times L_{Boundary}
    \label{eq:combined_loss}
\end{equation}
This particular  arrangement of the weights is chosen as a previous study has shown that it yields high accuracy in femur segmentation \cite{medvissuite}. 

\subsection{Training Details}
XAG-Net is trained using the Adam optimizer. The hyperparameters are set as follows: 
\begin{itemize}
    \item Epochs: 100
    \item Initial learning rate: $1\times10^{-4}$
    \item Batch size: 4
\end{itemize}
The model with the lowest validation loss is utilized for evaluation and comparison. XAG-Net was trained on an NVIDIA A100 GPU from Google Colab.

\section{Experimental Design}

\subsection{Dataset}
Our dataset consists of 4,802 axial MRI slices of the femur; each saved as a cropped PNG image of $360 \times 160$ pixels. A matching set of 4,802 binary mask images, drawn slice‑by‑slice by experienced musculoskeletal annotators, serves as voxel‑level ground truth for the bone outline.

All images are collected from ten adolescent subjects labeled TD01 through TD10. We utilize axial MRI femur scans from eight patients (TD01 to TD08) for the training dataset. Within each axial stack, the femur's proximal, shaft, and distal portions are all represented. Due to the shaft region being naturally over‑represented in a long‑bone scan, several shaft‑dominant slices are withdrawn from the particular training series to equalize the proportions of the main anatomies. After the process, the training corpus consists of 3,761 slices. For testing, 1,041 axial MRI slices of the femur from patients TD09 and TD10 are used. 

\subsection{Evaluation Metrics}
The three main metrics utilized to compare XAG-Net and other baseline models are the Dice similarity coefficient (DSC), intersection over union (IoU), and Hausdorff distance 95\% (HD95) \cite{HD95}. Both DSC and IoU scores measure the segmentation accuracy by comparing the predicted mask against the ground truth mask as values between 0 and 1, with values closer to 1 signifying better segmentation accuracy. The difference between the two metrics is that DSC assigns double the weight to the intersection, while IoU gives equal weight to intersection and union. For evaluation, the model outputs are binarized by applying a threshold of 0.5 before computing the DSC and IoU scores. The HD95 metric is calculated as the 95th percentile of the minimum distances between the prediction and ground truth boundaries in a nanmean approach. For HD95, values closer to 0 indicate better alignment of boundaries. For all these metrics, the overall score is calculated by computing the mean value across all evaluated slices. 

\subsection{Comparison}
We compare the DSC score, IoU score, and the HD95 performance of XAG-Net against those of U-Net \cite{ronneberger2015unet}, Attention U-Net \cite{oktay2018attentionunetlearninglook}, 2.5D U-Net \cite{ZHANG2022102088}, CSAM \cite{Yu2024CSAM}, and 3D U-Net \cite{Cicek2016_3DUNet}. All models are tested on axial femur MRI dataset consisting of 3,761 slices from patients 1 through 8 (TD01 through TD08) on 100 epochs. Evaluation of the models is done with 1,041 axial femur MRI scans from patients 9 and 10 (TD09 and TD10). We divide the evaluation into four parts. (i) A full-scan evaluation, which includes the entire femur structures from TD09 and TD10 as well as axial slices part of the scan immediately above the femur to test for false positives; (ii) Evaluation on the proximal femur; (iii) Evaluation on the femoral shaft; and (iv) Evaluation on the distal femur. This evaluation approach allows us to measure not only how XAG-Net would perform in a clinical setting where clinicians need to segment femur structures from full-scans, but also whether there are clear advantages or disadvantages of utilizing XAG-Net when evaluating the three major anatomical regions of the femur. 

\section{Experimental Results}
\subsection{Performance on Femur Segmentation}
\begin{figure}[t] \centering \includegraphics[width=0.65\linewidth]{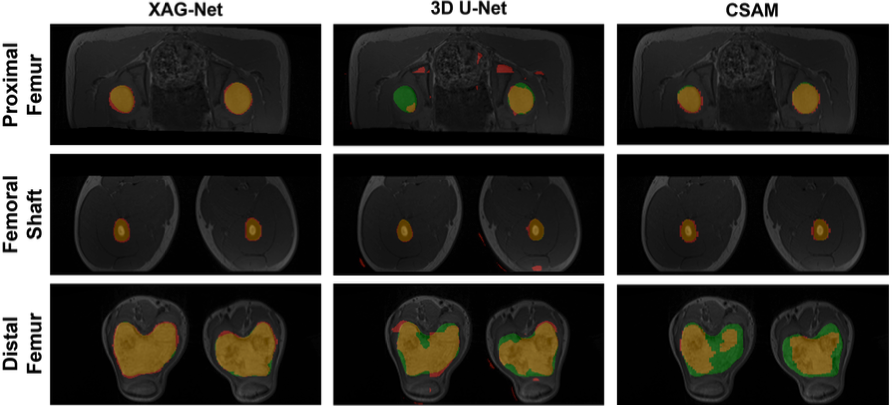} \caption{Visualization of segmentation output of XAG-Net, 3D U-Net, and CSAM. The green region represents the ground truth mask region, the red region represents the predicted mask region, and the yellow region represents the overlaid mask region.} \label{fig:2d_2.5d_3d} \end{figure}

\begin{table*}[t]
\caption{Full-Scan Evaluation Results of Different Models}
\begin{center}
\renewcommand{\arraystretch}{1.2}
\begin{tabular}{|c|c|c|c|c|c|}
\hline
\textbf{Model} & \textbf{DSC $\uparrow$} & \textbf{IoU} & \textbf{HD95} & \textbf{Parameter Count} & \textbf{FLOPs} \\
\hline
3D U-Net & 0.8307 & 0.7678 & 7.24 px & 52,965,074 & 325,250,285,568 \\
\hline
U-Net & 0.9335 & 0.8853 & 1.13 px & \textbf{18,950,593} & \textbf{62,760,240,384} \\
\hline
Attention U-Net & 0.9382 & 0.8907 & 1.34 px & 19,649,797 & 67,102,480,640 \\
\hline
2.5D U-Net & 0.9436 & 0.9061 & \textbf{0.92 px} & 18,956,865 & 63,582,323,968 \\
\hline
CSAM & 0.9466 & 0.8960 & 2.11 px & 34,720,653 & 327,336,504,356 \\
\hline
\textbf{XAG-Net} & \textbf{0.9535} & \textbf{0.9160} & \textbf{0.92 px} & 23,138,641 & 89,465,067,776 \\
\hline
\end{tabular}
\label{tab:fullscan_evaluation}
\end{center}
\end{table*}

\begin{table*}[t]
\caption{Regional Evaluation Results on Proximal, Shaft, and Distal Femur Regions}
\vspace{0.6em}

\renewcommand{\arraystretch}{1.65}

\begin{minipage}{0.47\textwidth}
\centering
\textbf{Proximal Femur Evaluation}\\[0.4em]
\begin{tabular}{|c|c|c|c|}
\hline
\textbf{Model} & \textbf{DSC $\uparrow$} & \textbf{IoU} & \textbf{HD95} \\
\hline
3D U-Net & 0.7664 & 0.6442 & 18.60 px \\
\hline
CSAM & 0.8880 & 0.8105 & 8.03 px \\
\hline
\makecell{Attention\\U-Net} & 0.8925 & 0.8138 & 3.85 px \\
\hline
U-Net & 0.8972 & 0.8195 & \textbf{3.56 px} \\
\hline
\makecell{2.5D\\U-Net} & 0.9126 & 0.8476 & 4.68 px \\
\hline
\textbf{XAG-Net} & \textbf{0.9197} & \textbf{0.8543} & 3.61 px \\
\hline
\end{tabular}
\end{minipage}
\hfill
\begin{minipage}{0.47\textwidth}
\centering
\textbf{Femoral Shaft Evaluation}\\[0.4em]
\begin{tabular}{|c|c|c|c|}
\hline
\textbf{Model} & \textbf{DSC $\uparrow$} & \textbf{IoU} & \textbf{HD95} \\
\hline
3D U-Net & 0.9337 & 0.8772 & 4.00 px \\
\hline
\makecell{Attention\\U-Net} & 0.9468 & 0.8999 & 0.89 px \\
\hline
U-Net & 0.9476 & 0.9010 & 0.44 px \\
\hline
CSAM & 0.9611 & 0.9252 & 0.83 px \\
\hline
\makecell{2.5D\\U-Net} & \textbf{0.9626} & \textbf{0.9280} & \textbf{0.12 px} \\
\hline
\textbf{XAG-Net} & 0.9624 & 0.9277 & 0.22 px \\
\hline
\end{tabular}
\end{minipage}

\vspace{1em}

\begin{center}
\begin{minipage}{0.6\textwidth}
\centering
\textbf{Distal Femur Evaluation}\\[0.4em]
\begin{tabular}{|c|c|c|c|}
\hline
\textbf{Model} & \textbf{DSC $\uparrow$} & \textbf{IoU} & \textbf{HD95} \\
\hline
3D U-Net & 0.7647 & 0.6757 & 15.86 px \\
\hline
CSAM & 0.8909 & 0.8198 & 3.98 px \\
\hline
U-Net & 0.9091 & 0.8437 & 3.08 px \\
\hline
\makecell{2.5D\\U-Net} & 0.9162 & 0.8600 & 2.23 px \\
\hline
\makecell{Attention\\U-Net} & \textbf{0.9256} & \textbf{0.8701} & \textbf{1.70 px} \\
\hline
\textbf{XAG-Net} & 0.9235 & 0.8682 & 2.42 px \\
\hline
\end{tabular}
\end{minipage}
\end{center}

\vspace{0.2cm}
\noindent\footnotesize{Footnote: DSC and IoU are overall scores computed across all slices. Models are sorted in ascending DSC ($\uparrow$) order, with XAG-Net placed last.}

\label{tab:regional_evaluation}
\end{table*}

\textit{1) Full-Scan Dataset:} The results of the full-scan evaluation are presented in Table~\ref{tab:fullscan_evaluation}. XAG-Net surpasses all baseline models across all segmentation metrics - DSC (0.9535), IoU (0.9160), and HD95 (0.92 px) - while remaining computationally efficient. Compared to 3D U-Net, we see that XAG-Net requires 56.3\% fewer parameters and shows a 72\% reduction in the number of FLOPs, while achieving a 14.78\% increase in DSC and a 19.30\% increase in IoU. XAG-Net also outperforms CSAM in segmentation accuracy, while requiring 33.35\% fewer parameters and 72.67\% fewer FLOPs. CSAM achieves a higher DSC than 2.5D U-Net but underperforms in both IoU and HD95. 2.5D U-Net yields higher DSC and IoU scores than the baseline 2D models, while incurring a lower computational cost than Attention U-Net. Attention U-Net demonstrates higher segmentation accuracy than U-Net, with a modest increase in computational cost. 3D U-Net performs the worst in segmentation accuracy and incurs the highest computational cost out of all baseline models. The average inference time for XAG-Net is 250.89 ms, which is 69.01 ms longer than that of the baseline U-Net (181.88 ms).

\textit{2) Regional Femur Datasets:}
From the results in the proximal femur section of the regional evaluation in Table~\ref{tab:regional_evaluation}, we find that XAG-Net achieves the highest DSC and IoU scores. On the femoral shaft dataset, XAG-Net also outperforms all except 2.5D U-Net by a minuscule margin, i.e., 0.0002 in DSC. Results are similar for the distal femur, where XAG-Net outperforms all baseline models except Attention U-Net by a minuscule margin, i.e., 0.0021 in DSC. Notably, the distal femur is the only region where a 2D model, i.e., Attention U-Net, outperforms all 2.5D baseline models. From this evaluation, we observe that XAG-Net consistently outperforms most baseline models across all femoral regions, with the proximal femur showing the largest gain.

\subsection{Ablation Study}
\begin{table}[htbp]
\caption{Ablation Study Results}
\begin{center}
\renewcommand{\arraystretch}{1.2}
\begin{tabular}{|c|c|c|c|c|}
\hline
\textbf{Input CSA} & \textbf{Skip CSA} & \textbf{Skip AG} & \textbf{DSC} & \textbf{IoU} \\
\hline
& & & 0.9436 & 0.9061 \\ 
\hline
\checkmark & & & 0.9449 & 0.9076 \\ 
\hline
& \checkmark & & 0.9520 & 0.9148 \\ 
\hline
& & \checkmark & 0.9341 & 0.8955 \\ 
\hline
\checkmark & \checkmark & & 0.9518 & 0.9148 \\ 
\hline
\checkmark & & \checkmark & 0.9406 & 0.9020 \\ 
\hline
& \checkmark & \checkmark & 0.9504 & 0.9139 \\ 
\hline
\checkmark & \checkmark & \checkmark & \textbf{0.9535} & \textbf{0.9160} \\ 
\hline
\end{tabular}
\label{tab:ablation_study}
\end{center}
\end{table}
The ablation study evaluates the significance of the main components of the proposed XAG-Net architecture: the input CSA module, CSA modules in skip connections, and AG blocks in skip connections. The ablation models are created as different combinations of these components. For training, axial femur MRI datasets TD01 through TD08 are utilized, and for testing, those from TD09 and TD10 are utilized.

From the results presented in the ablation study in Table~\ref{tab:ablation_study}, CSA modules in skip connections contribute most to the performance improvement. On the other hand, AG blocks in skip connections do not deliver any performance enhancement when used alone. However, when combined with CSA modules, their contribution becomes beneficial - likely because CSA modules provide critical inter-slice context that mitigates the limitations of AG while allowing any valuable intra-slice features to be effectively leveraged during feature concatenation. XAG-Net, comprising of Input CSA, Skip CSA, and Skip AG, outperforms all other models, which illustrates the enhanced segmentation accuracy achieved from the amalgamation of these components and the efficacy of their placement within the architecture. 

\section{Discussion}
The experimental results show that XAG-Net consistently surpasses baseline 2D, 2.5D, and 3D models in femur MRI segmentation. XAG-Net's superior performance stems from its unique integration of pixel-wise CSA and skip-level AG blocks, enabling fine-grained inter-slice context modeling and intra-slice refinement not achieved by previous 2.5D or 3D approaches. This demonstrates that the amalgamation of the input CSA module, skip CSA module, and skip AG blocks is critical for providing inter- and intra-slice contexts to segment the target structure across all femur regions. The 2.5D approach within the U-Net architecture enhances segmentation accuracy more than AG blocks. Therefore, when improving segmentation accuracy of femur MRI scans, preserving partial volumetric context via a 2.5D structure is more crucial than suppressing irrelevant background via AG blocks.

From Table \ref{tab:fullscan_evaluation}, XAG-Net outperforms 2.5D U-Net and CSAM in segmentation accuracy while only incurring a modest computational cost increase relative to 2.5D U-Net. This demonstrates that the main components of XAG-Net effectively and efficiently enhance the partial volumetric context by incorporating inter-slice information from neighboring slices and focusing on salient features. When comparing the segmentation performance of XAG-Net and CSAM, it is essential to note that CSAM was developed to mainly segment large, amorphous, or soft-tissue structures like the prostate or placenta \cite{Yu2024CSAM}. Segmenting soft-tissue structures may pose a different challenge than segmenting rigid cortical boundaries with minor anatomical variations like the femur bone. XAG-Net's pixel-wise softmax attention approach across just three slices may preserve local variations better than global modeling across all slices when working with rigid cortical boundaries. In addition, 3D U-Net's underperformance can be attributed to 3D U-Net models' high complexity and propensity to overfit when trained on smaller datasets \cite{ZHANG2022102088},\cite{haque2019semanticsegmentationthighmuscle}. 

From Table \ref{tab:regional_evaluation}, XAG-Net outperforms every baseline model on the proximal femur dataset. The proximal femur has more complex 3D shapes than the femoral shaft or distal femur. Thus, successfully modeling volumetric context is crucial in accurately segmenting the proximal femur, which XAG-Net would excel in due to its CSA module. The femoral shaft region is a relatively uniform structure, meaning that the volumetric context provided in 2.5D U-Net is enough to segment this region accurately. The distal femur has a much lighter local texture variation and requires a detailed edge delineation, which explains Attention U-Net's ideal performance. The consistently strong performance of XAG-Net across all femur regions demonstrates its ability to balance segmentation accuracy with computational efficiency.

One limitation of this study is that the models are trained on a single institutional dataset of 3,761 axial femur MRI slices. Testing XAG-Net on a public dataset or external institutional data could strengthen the model's generalizability. Although XAG-Net remains computationally efficient compared to the full-volumetric 3D U-Net and a more complex 2.5D architecture like CSAM, it still incurs a higher computational cost than 2.5D U-Net and 2D models. Therefore, optimizing the components of XAG-Net to make it more computationally efficient - such as through model pruning - while preserving or improving the model's segmentation accuracy could be a valuable avenue of future work.

\section{Conclusion}
In this study, we introduced XAG-Net, a novel 2.5D U-Net model for improved femur MRI segmentation with cross-slice attention (CSA) and skip Attention Gating (AG) mechanisms. Evaluation of XAG-Net against baseline 2D, 2.5D, and 3D models showed that XAG-Net outperforms these baselines in segmentation accuracy while remaining computationally efficient. XAG-Net outperformed 3D U-Net on segmentation accuracy and computational efficiency, indicating its robustness in low-data regimes where fully volumetric models like 3D U-Nets often struggle. An extensive ablation study confirmed the key role of the CSA module, especially in skip connections of the architecture. While XAG-Net introduced a modest increase in computational cost compared to 2D baseline models and 2.5D U-Net, it balances segmentation accuracy and model complexity. Possible future work includes external validation on new datasets and architectural optimizations to further reduce computational overhead.
\bibliographystyle{IEEEtran} 

\begin{thebibliography}{10}
\providecommand{\url}[1]{#1}
\csname url@samestyle\endcsname
\providecommand{\newblock}{\relax}
\providecommand{\bibinfo}[2]{#2}
\providecommand{\BIBentrySTDinterwordspacing}{\spaceskip=0pt\relax}
\providecommand{\BIBentryALTinterwordstretchfactor}{4}
\providecommand{\BIBentryALTinterwordspacing}{\spaceskip=\fontdimen2\font plus
\BIBentryALTinterwordstretchfactor\fontdimen3\font minus \fontdimen4\font\relax}
\providecommand{\BIBforeignlanguage}[2]{{%
\expandafter\ifx\csname l@#1\endcsname\relax
\typeout{** WARNING: IEEEtran.bst: No hyphenation pattern has been}%
\typeout{** loaded for the language `#1'. Using the pattern for}%
\typeout{** the default language instead.}%
\else
\language=\csname l@#1\endcsname
\fi
#2}}
\providecommand{\BIBdecl}{\relax}
\BIBdecl

\bibitem{Xie2025}
\BIBentryALTinterwordspacing
W.~Xie, P.~Chen, Z.~Li, X.~Wang, C.~Wang, L.~Zhang, W.~Wu, J.~Xiang, Y.~Wang, and D.~Zhong, ``A two-stage deep learning network for automated femoral segmentation in bilateral lower limb ct scans,'' \emph{Scientific Reports}, vol.~15, no.~1, p. 9198, 2025. [Online]. Available: \url{https://doi.org/10.1038/s41598-025-94180-1}
\BIBentrySTDinterwordspacing

\bibitem{FOURNEL2021102213}
\BIBentryALTinterwordspacing
J.~Fournel, A.~Bartoli, D.~Bendahan, M.~Guye, M.~Bernard, E.~Rauseo, M.~Y. Khanji, S.~E. Petersen, A.~Jacquier, and B.~Ghattas, ``Medical image segmentation automatic quality control: A multi-dimensional approach,'' \emph{Medical Image Analysis}, vol.~74, p. 102213, 2021. [Online]. Available: \url{https://www.sciencedirect.com/science/article/pii/S1361841521002589}
\BIBentrySTDinterwordspacing

\bibitem{ronneberger2015unet}
O.~Ronneberger, P.~Fischer, and T.~Brox, ``U-net: Convolutional networks for biomedical image segmentation,'' in \emph{Medical Image Computing and Computer-Assisted Intervention -- MICCAI 2015}, N.~Navab, J.~Hornegger, W.~M. Wells, and A.~F. Frangi, Eds.\hskip 1em plus 0.5em minus 0.4em\relax Cham: Springer International Publishing, 2015, pp. 234--241.

\bibitem{ZHANG2022102088}
\BIBentryALTinterwordspacing
Y.~Zhang, Q.~Liao, L.~Ding, and J.~Zhang, ``Bridging 2d and 3d segmentation networks for computation-efficient volumetric medical image segmentation: An empirical study of 2.5d solutions,'' \emph{Computerized Medical Imaging and Graphics}, vol.~99, p. 102088, 2022. [Online]. Available: \url{https://www.sciencedirect.com/science/article/pii/S0895611122000611}
\BIBentrySTDinterwordspacing

\bibitem{Cicek2016_3DUNet}
O.~\c{C}i\c{c}ek, A.~Abdulkadir, S.~S. Lienkamp, T.~Brox, and O.~Ronneberger, ``3d u-net: Learning dense volumetric segmentation from sparse annotation,'' in \emph{Medical Image Computing and Computer-Assisted Intervention -- MICCAI 2016}, S.~Ourselin, L.~Joskowicz, M.~R. Sabuncu, G.~Unal, and W.~Wells, Eds.\hskip 1em plus 0.5em minus 0.4em\relax Cham: Springer International Publishing, 2016, pp. 424--432.

\bibitem{Nikzad2024CSANet}
\BIBentryALTinterwordspacing
A.~Kumar, H.~Jiang, M.~Imran, C.~Valdes, G.~Leon, D.~Kang, P.~Nataraj, Y.~Zhou, M.~D. Weiss, and W.~Shao, ``A flexible 2.5d medical image segmentation approach with in-slice and cross-slice attention,'' \emph{Computers in Biology and Medicine}, vol. 182, p. 109173, 2024. [Online]. Available: \url{https://www.sciencedirect.com/science/article/pii/S0010482524012587}
\BIBentrySTDinterwordspacing

\bibitem{Deniz_2018}
\BIBentryALTinterwordspacing
C.~M. Deniz, S.~Xiang, R.~S. Hallyburton, A.~Welbeck, J.~S. Babb, S.~Honig, K.~Cho, and G.~Chang, ``Segmentation of the proximal femur from mr images using deep convolutional neural networks,'' \emph{Scientific Reports}, vol.~8, no.~1, Nov. 2018. [Online]. Available: \url{http://dx.doi.org/10.1038/s41598-018-34817-6}
\BIBentrySTDinterwordspacing

\bibitem{liu2025performanceanalysisdeeplearning}
\BIBentryALTinterwordspacing
M.~Liu, Y.~Chen, A.~Tian, X.~Wu, M.~Shen, T.~Gong, and J.~Lee, ``Performance analysis of deep learning models for femur segmentation in mri scan,'' 2025. [Online]. Available: \url{https://arxiv.org/abs/2504.04066}
\BIBentrySTDinterwordspacing

\bibitem{Yu2024CSAM}
Y.~H. AL, H.~Zheng, K.~Zhao, X.~Du, K.~Pang, Q.~Miao, S.~S. Raman, D.~Terzopoulos, and K.~Sung, ``{CSAM: A 2.5D Cross-Slice Attention Module for Anisotropic Volumetric Medical Image Segmentation},'' in \emph{Proceedings of the IEEE/CVF Winter Conference on Applications of Computer Vision (WACV)}, 2024, pp. 5911--5920, pMCID: PMC11349312, Epub 2024 Apr 9.

\bibitem{oktay2018attentionunetlearninglook}
\BIBentryALTinterwordspacing
O.~Oktay, J.~Schlemper, L.~L. Folgoc, M.~Lee, M.~Heinrich, K.~Misawa, K.~Mori, S.~McDonagh, N.~Y. Hammerla, B.~Kainz, B.~Glocker, and D.~Rueckert, ``Attention u-net: Learning where to look for the pancreas,'' 2018. [Online]. Available: \url{https://arxiv.org/abs/1804.03999}
\BIBentrySTDinterwordspacing

\bibitem{2.5D_UNet}
K.~Hu, C.~Liu, X.~Yu, J.~Zhang, Y.~He, and H.~Zhu, ``A 2.5d cancer segmentation for mri images based on u-net,'' in \emph{2018 5th International Conference on Information Science and Control Engineering (ICISCE)}, 2018, pp. 6--10.

\bibitem{Zhang_2019}
H.~Zhang, J.~Zhang, Q.~Zhang, J.~Kim, S.~Zhang, S.~A. Gauthier, P.~Spincemaille, T.~D. Nguyen, M.~Sabuncu, and Y.~Wang, ``Rsanet: Recurrent slice-wise attention network for multiple sclerosis lesion segmentation,'' in \emph{Proceedings of the International Conference on Medical Image Computing and Computer-Assisted Intervention (MICCAI)}, Cham, 2019, pp. 411--419.

\bibitem{medvissuite}
\BIBentryALTinterwordspacing
{Liu, Mengyuan}, {Zhang, Di}, {Chen, Yixiao}, {Gong, Tianchou}, {Kainz, Hans}, {Song, Seungmoon}, and {Lee, Jeongkyu}, ``Medvis suite: A framework for mri visualization and u-net-based bone segmentation with in-depth evaluation,'' \emph{BIO Web Conf.}, vol. 163, p. 04001, 2025. [Online]. Available: \url{https://doi.org/10.1051/bioconf/202516304001}
\BIBentrySTDinterwordspacing

\bibitem{HD95}
\BIBentryALTinterwordspacing
A.~Celaya, B.~Riviere, and D.~Fuentes, ``A generalized surface loss for reducing the hausdorff distance in medical imaging segmentation,'' 2024. [Online]. Available: \url{https://arxiv.org/abs/2302.03868}
\BIBentrySTDinterwordspacing

\bibitem{haque2019semanticsegmentationthighmuscle}
\BIBentryALTinterwordspacing
H.~Haque, M.~Hashimoto, N.~Uetake, and M.~Jinzaki, ``Semantic segmentation of thigh muscle using 2.5d deep learning network trained with limited datasets,'' 2019. [Online]. Available: \url{https://arxiv.org/abs/1911.09249}
\BIBentrySTDinterwordspacing

\end{thebibliography}

\end{document}